\documentclass[runningheads]{llncs}

\usepackage[T1]{fontenc}
\usepackage{graphicx}
\usepackage{booktabs}
\usepackage{amsmath,amssymb}
\usepackage{microtype}
\usepackage{xcolor}
\usepackage{url}
\usepackage{placeins}
\usepackage{bbding}
\usepackage[hidelinks]{hyperref}
\hypersetup{
  pdftitle={Dynamic K-Head: Difficulty-Supervised Budget Selection for Continuous Latent Reasoning},
  pdfauthor={Ying Li, Yuqi Chen, Sixuan Li, Yunfeng Cai},
  pdfkeywords={Latent reasoning, Continuous chain-of-thought, Adaptive computation, Efficient reasoning, Knowledge distillation}
}
\definecolor{dkblue}{HTML}{4C78A8}
\definecolor{dkred}{HTML}{E45756}
\definecolor{dkgreen}{HTML}{59A14F}
\definecolor{dkorange}{HTML}{F28E2B}
\definecolor{dkgray}{HTML}{666666}

\newcommand{\method}{Dynamic K-Head}

\newcommand{\kset}{\{1,2,3,4,6,8\}}
\newcommand{\bottok}{\texttt{BOT}}
\newcommand{\eottok}{\texttt{EOT}}

\begin{document}

\title{Dynamic K-Head: Difficulty-Supervised Budget Selection for Continuous Latent Reasoning}
\titlerunning{Dynamic K-Head for Continuous Latent Reasoning}

\author{Ying Li\inst{1,2} \and
Yuqi Chen\inst{3(\text{\Envelope})} \and
Sixuan Li\inst{4} \and
Yunfeng Cai\inst{1(\text{\Envelope})}}
\authorrunning{Y. Li et al.}
\institute{Beijing Institute of Mathematical Sciences and Applications, Beijing, China\\
\email{caiyunfeng@bimsa.cn} \and
Tsinghua University, Beijing, China\\
\email{liying0928@tsinghua.edu.cn} \and
The University of Hong Kong, Hong Kong, China\\
\email{chenyuqi@hku.hk} \and
Peking University, Beijing, China\\
\email{sxli@pku.edu.cn}}

\maketitle

\begin{abstract}
Chain-of-thought (CoT) improves reasoning but incurs substantial autoregressive rationale cost. Continuous CoT moves intermediate computation into recurrent hidden states, yet typically uses a fixed number of updates even though the best budget varies across questions and accuracy need not improve monotonically with $K$. We introduce \method{}, a question-conditioned, pre-rollout budget selector. It is jointly trained with the reasoner under selected budgets, using difficulty supervision based on scores assigned from the question alone. We also show that latent-update count can misrepresent end-to-end cost because a short latent rollout can shift computation into a longer explicit continuation. We therefore count latent updates, implementation-specific special tokens, and all decoded text. Instantiated with CoDI on Llama 3.2 1B and GSM8K, \method{} reaches 54.44\% exact-match accuracy at 12.03 average reasoning/output tokens, compared with 52.31\% at 15.21 tokens for fixed-$K=6$ CoDI---a 2.12-point gain with 20.9\% lower token cost. Across three out-of-distribution arithmetic datasets, it reduces cost by 19.3--25.7\% while improving accuracy by 0.30--2.22 points. These results demonstrate that difficulty-supervised budget selection can improve both accuracy and efficiency, while stable budget selection remains a central challenge.

\end{abstract}

\section{Introduction}

Chain-of-thought (CoT) reasoning improves language-model performance by making intermediate computation explicit \cite{wei2022chain,kojima2022zero,wang2023selfconsistency,cobbe2021gsm8k}, but the resulting rationales can dominate decoding cost. Recent latent-reasoning methods instead feed hidden states back into the model as continuous thoughts. Coconut trains models to reason in a continuous latent space \cite{hao2024coconut}, while CoDI jointly trains explicit and implicit paths through self-distillation \cite{shen2025codi}. These methods reduce verbal output, yet most expose a fixed inference hyperparameter: the number $K$ of latent updates.

\begin{figure}[t]
\centering
\includegraphics[width=\textwidth]{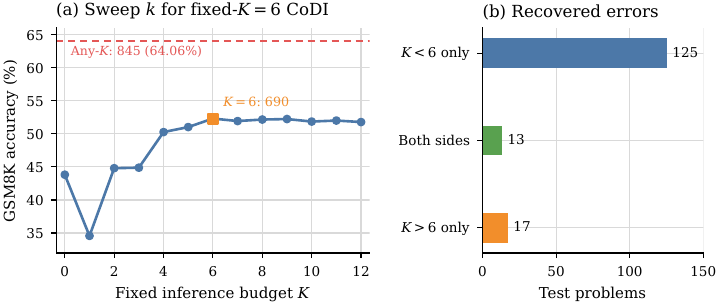}
\caption{Fixed-budget accuracy is non-monotonic and conceals instance-level recoveries. Panel (a) sweeps inference budget $k$ for CoDI trained with fixed $K=6$. Panel (b) partitions the 155 $K=6$ errors recovered by another swept budget: 125 are recovered only below six, 17 only above six, and 13 on both sides. Thus the smaller- and larger-budget recovery totals are 138 and 30, respectively, with the 13 ``both sides'' problems included in each.}
\label{fig:oracle}
\end{figure}

Figure~\ref{fig:oracle} illustrates why a single $K$ is an awkward resource-allocation scheme. Easy questions may need little additional computation, whereas hard questions may benefit from more. More importantly, recurrent latent computation is not guaranteed to improve predictions monotonically. A later hidden state can overwrite a useful partial solution, amplify a shortcut, or move the final decoder into a less favorable region. With Llama 3.2 1B CoDI, $K=6$ solves 690 of 1,319 GSM8K test problems. Sweeping $K=0,\ldots,12$ with the same checkpoint shows that 845 problems are solved by at least one budget. Of the 629 questions missed at $K=6$, 138 are recovered by some smaller budget and 30 by some larger budget; these recovery sets overlap. Thus adaptive budget selection is not merely a rule of ``more steps for harder questions.'' It requires an instance-conditioned choice among a non-monotonic family of computations.

Our \method{} attaches a linear head to the pre-rollout question representation and selects the number of latent states actually executed, jointly exposing the K-head and reasoner to variable computation. The contrastive objective implementing difficulty supervision pulls together budget representations of questions with similar difficulty and separates dissimilar ones. Training begins with a fixed-budget warm-up and gradually transfers budget selection to the K-head; inference chooses its highest-scoring budget. The interface requires only a controllable recurrent rollout and is instantiated with CoDI in our experiments.

Relative to CoDI, \method{} improves accuracy while reducing reasoning/output token cost. With $\lambda_K=0.01$, it reaches 54.51\% test accuracy at 13.33 average tokens, a 2.20-point accuracy gain with 12.4\% fewer tokens than CoDI's 52.31\% accuracy at 15.21 tokens. With $\lambda_K=0.03$, it reaches 54.44\% accuracy at 12.03 tokens, retaining a 2.12-point gain while reducing token cost by 20.9\%.

Our contributions are fourfold:
\begin{itemize}
    \item We show that fixed latent budgets conceal non-monotonic instance-level variation: problems missed at one budget can be recovered by either smaller or larger budgets, motivating adaptive budget selection beyond a ``more computation for harder questions'' rule.
    \item We show that latent-token count alone can provide an imprecise estimate of end-to-end cost: a shorter latent rollout can shift reasoning into a longer explicit continuation. Our end-to-end accounting includes latent updates, implementation-specific special tokens, and all subsequently decoded text, and is applied uniformly to enable fair token-cost comparisons across different experimental settings.
    \item We introduce \method{}, a pre-rollout, question-conditioned budget selector for recurrent latent reasoners, jointly trained under selected budgets with difficulty supervision, without direct budget labels or a monotonic difficulty-to-budget mapping.
    \item \method{} improves CoDI by 2.12 accuracy points while reducing reasoning/output token cost by 20.9\%; without difficulty supervision, budget selection is less accurate and more expensive.
\end{itemize}

\section{Related Work}

\subsection{Explicit and Latent Reasoning}

Explicit CoT exposes intermediate natural-language steps and has become a standard tool for mathematical reasoning \cite{wei2022chain,cobbe2021gsm8k}. Zero-shot CoT broadens its use without exemplars \cite{kojima2022zero}, while self-consistency samples multiple explicit traces and aggregates their answers \cite{wang2023selfconsistency}. These techniques improve reasoning but retain, or multiply, autoregressive rationale cost. Pause-token methods insert learned computation slots before an answer \cite{goyal2024pause}, while Quiet-STaR trains language models to generate internal rationales before predicting future text \cite{zelikman2024quietstar}. Both lines suggest that inference-time computation can be separated from user-visible output.

Continuous latent reasoning removes the requirement that every intermediate state correspond to a vocabulary item. Earlier implicit-CoT work distills explicit reasoning into layerwise hidden computation \cite{deng2023implicit} or progressively removes rationale steps during fine-tuning \cite{deng2024internalize}. SoftCoT instead uses a lightweight assistant to produce continuous soft thoughts for an unmodified target LLM \cite{xu2025softcot}. Coconut recurrently feeds the last hidden state back as the next input embedding and observes search-like behavior on logical tasks \cite{hao2024coconut}. CoDI uses one shared model as teacher and student: an explicit CoT path supplies hidden-state targets to a latent path, aligning the two at answer positions \cite{shen2025codi}. PCCoT applies Jacobi-style parallel updates to latent tokens to improve training and inference throughput \cite{wu2025pccot}. These approaches motivate compressed reasoning, but do not learn a pre-rollout, question-conditioned budget selector for recurrent latent computation.

Recent analyses suggest that latent computation is not a sequence of interchangeable depth increments: causal interventions reveal staged, non-local functionality across latent steps, while mechanistic probes find complete intermediate-state sequences on shorter-hop tasks but partial, late-intermediate pathways on longer-hop tasks \cite{li2026dynamics,liang2026mechanistic}. This perspective helps explain our non-monotonic $K$ sweeps and motivates adaptive budget selection rather than monotonic depth allocation.

\subsection{Adaptive Computation}

Adaptive computation has a long history in recurrent and universal architectures. Adaptive Computation Time learns a halting probability with a ponder cost \cite{graves2016act}; PonderNet places a probabilistic model over computation depth \cite{banino2021pondernet}; and the Universal Transformer repeatedly refines token representations with optional dynamic halting \cite{dehghani2019universal}. Depth-Adaptive Transformer and CALM exit from intermediate Transformer layers at sequence or generation-token granularity \cite{elbayad2020depth,schuster2022calm}, whereas Mixture-of-Depths selectively processes token positions at each layer under a fixed total capacity \cite{raposo2024mixture}. These methods establish the value of conditional compute, but adapt layer execution rather than selecting the number of recurrent continuous thoughts before answer decoding.

For latent language-model reasoning, Adaptive Latent Reasoning learns online stop/continue decisions through post-SFT reinforcement learning, whereas AdaAnchor applies a hand-designed stability rule while training anchors and LoRA modules around a frozen dense backbone \cite{ning2025adaptive,sheshanarayana2026adaanchor}. \method{} instead selects a latent budget for each question before rollout. Two distinctions are central. First, the K-head and LoRA-adapted reasoner are jointly trained under the budgets actually executed. Second, $K$ is not an ordered refinement axis: under non-monotonic dynamics, a halting mechanism can only stop or append another update, and continuing may irreversibly drift from a useful state. We therefore treat budgets as distinct choices and apply difficulty supervision without optimal-$K$ labels or a monotonic mapping.

\subsection{Difficulty Supervision}

Example difficulty has long been used to order training curricula \cite{bengio2009curriculum}, and model training dynamics can expose easy, ambiguous, and hard data regions \cite{swayamdipta2020dataset}. Difficulty is not directly observed in standard reasoning datasets, however, and need not vary monotonically with the best latent budget. Teacher-model judgments can provide weak supervision, but may reflect the annotator's priors rather than the student's computation. We therefore avoid regressing $K$ onto an assumed target. Instead, a contrastive objective uses only relative neighborhoods in the scalar difficulty scores. This design borrows the principle of pulling related examples together and separating unrelated ones from self-supervised and supervised contrastive learning \cite{chen2020simclr,khosla2020supervised}, but uses neither augmentation identity nor discrete class or budget labels.

\section{Background and Problem Formulation}

Standard CoT externalizes intermediate reasoning as a sequence of natural-language tokens before producing the answer, so every rationale token incurs an autoregressive decoding step. Latent CoT instead carries out this intermediate computation in continuous hidden-state space \cite{hao2024coconut,shen2025codi}. Given a question $q$, the model first prefills the question. At each latent update, it takes the hidden state at the current final position, converts it into the continuous input representation for the next position, and feeds that representation back into the model without selecting a vocabulary token. The resulting final-position hidden state is then used for the next update, forming a recurrent latent rollout.

We denote the number of latent updates by $K$. After exactly $K$ updates, the model leaves the latent rollout and generates the answer $y$ as ordinary vocabulary tokens with the standard autoregressive decoder, conditioned on both the question and the accumulated latent states. A fixed-budget latent reasoner uses the same number of updates $K$ for every question. Let $\mathcal{K}$ denote the allowed budgets and let $k$ index one candidate. We instead select
\begin{equation}
\widehat K(q)=\arg\max_{k\in\mathcal{K}}p_\phi(k\mid q)
\label{eq:adaptive-budget}
\end{equation}
before latent reasoning, and perform exactly $\widehat K(q)$ of the latent updates described above.

\section{Dynamic K-Head}
\label{sec:dynamic-k-head}

\subsection{Pre-Rollout Budget Selection}

Let $h_q$ be the question representation extracted before latent reasoning. A K-head $g_\phi$ maps $h_q$ to logits over an allowed budget set $\mathcal{K}$:
\begin{equation}
a = g_\phi(h_q), \qquad
p_\phi(k\mid q;\tau) = \operatorname{softmax}(a/\tau)_k.
\label{eq:khead}
\end{equation}
Here $\tau$ controls the sharpness of the budget distribution.

Training begins with an anchor-budget warm-up. Let $\rho$ denote the warm-up fraction and $K_a\in\mathcal{K}$ the anchor budget. During the first $\rho$ fraction of optimization, every example uses $K_a$ and the K-head objective is disabled. After warm-up, we define an anchor-mixture prior
\begin{equation}
p_0(k)=(1-\alpha)\frac{1}{|\mathcal{K}|}
+\alpha\mathbf{1}[k=K_a],
\end{equation}
where $\alpha$ controls the extra mass assigned to the anchor. Each example then independently draws its training-time executed budget according to
\begin{equation}
\begin{aligned}
\widehat K_i &\sim p_{\mathrm{train}}(\cdot\mid q_i),\\
p_{\mathrm{train}}(k\mid q_i)
&=(1-\epsilon)p_\phi(k\mid q_i;\tau)+\epsilon p_0(k).
\end{aligned}
\label{eq:sampling}
\end{equation}
Over the post-warm-up portion of training, $\tau$ is annealed from $\tau_{\mathrm{start}}$ to $\tau_{\mathrm{end}}$, while $\epsilon$ is annealed from $\epsilon_{\mathrm{start}}$ to $\epsilon_{\mathrm{end}}$. The concrete K-head architecture, budget set, and schedule values are specified in Section~5.

\begin{figure}[htbp]
\centering
\includegraphics{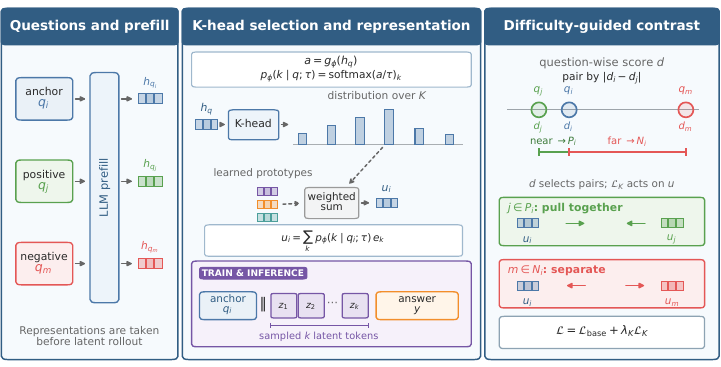}
\caption{Schematic of \method{} training and execution. Anchor, positive, and negative questions are prefilled by the LLM, and the K-head produces a distribution over latent budgets. Learned prototypes map this distribution to $u_i$, while per-question difficulty scores make nearby questions positive pairs and distant questions negative pairs. A sampled $K$ controls the illustrated shared train/inference path, where the anchor question is concatenated with exactly $K$ latent tokens before answer decoding.}
\label{fig:method}
\end{figure}

\subsection{Difficulty Supervision}

Each training question has a scalar difficulty score $d_i$ assigned by a teacher given only the question. To obtain a differentiable representation of the K-head output, we learn one prototype $e_k$ per candidate budget and form
\begin{equation}
u_i = \sum_{k\in\mathcal{K}}p_\phi(k\mid q_i;\tau)e_k.
\end{equation}
Within a batch $B$, examples are ranked for anchor $i$ by $|d_i-d_j|$. The $n_+$ nearest examples form the positive set $P_i$, and the $n_-$ farthest form the negative set $N_i$. Let $A_i=P_i\cup N_i$ denote the comparison set. Adapting the standard supervised contrastive objective \cite{khosla2020supervised} to these difficulty-defined sets, with cosine similarity $s$ and temperature $T$, we use
\begin{equation}
\mathcal{L}_{K}=-\frac{1}{|B|}\sum_{i\in B}\frac{1}{|P_i|}
\sum_{p\in P_i}\log
\frac{\exp(s(u_i,u_p)/T)}
{\sum_{a\in A_i}\exp(s(u_i,u_a)/T)}.
\end{equation}
The final objective is
\begin{equation}
\mathcal{L}=\mathcal{L}_{\mathrm{base}}+\lambda_K\mathcal{L}_{K}.
\label{eq:objective}
\end{equation}
Here $\mathcal{L}_{\mathrm{base}}$ is the original training objective of the corresponding fixed-budget latent reasoner. This difficulty supervision can therefore be added to different latent-reasoning methods without changing their method-specific base objectives. It encourages nearby difficulty scores to induce similar budget distributions without imposing a target $K$ or a monotonic difficulty-to-budget mapping.

\subsection{Token-Cost Accounting}

Prior latent-reasoning studies, for example CoDI and Adaptive Latent Reasoning, measure token cost primarily by latent reasoning length, thereby using the number of latent tokens or updates as a proxy for cost \cite{shen2025codi,ning2025adaptive}. This proxy can be imprecise because a model may continue reasoning in text after leaving latent mode: reducing $K$ may merely shift computation from latent updates to explicit autoregressive decoding rather than reduce the total cost. Appendix~\ref{app:continuation} provides a real GSM8K test example. The model selects only one latent update, $\widehat K=1$, but then generates 28 text tokens containing three explicit calculation steps and the final answer; together with the two special tokens in this CoDI execution, its total cost is $1+2+28=31$ tokens. Thus, a smaller latent budget does not necessarily imply a lower end-to-end reasoning/output cost. We therefore count the latent computation, the special tokens required by the latent-reasoning implementation, and all subsequently generated text. For a question $q$ executed with latent budget $K$, the cost is
\begin{equation}
C(q,K)=K+T_{\mathrm{special}}+T_{\mathrm{text}}(q,K),
\label{eq:cost}
\end{equation}
where $K$ counts the executed latent updates, $T_{\mathrm{special}}$ counts implementation-specific special tokens used for latent reasoning, such as \bottok{} and \eottok{}, and $T_{\mathrm{text}}(q,K)$ counts every token decoded after the latent rollout, including any explicit reasoning and the final answer.

\subsection{Deterministic Per-Example Inference}

At inference, prior mixing and stochastic sampling are removed. The K-head selects
\begin{equation}
\widehat K(q)=\arg\max_{k\in\mathcal{K}}p_\phi(k\mid q).
\end{equation}
The reasoner then executes exactly $\widehat K(q)$ latent updates before decoding the answer.

\subsection{Active-Then-Repack Execution}

Different examples in a batch may execute different budgets. The implementation maintains only rows that still require another latent update and removes a row from the active cache once it has reached its selected budget. It then concatenates each question with exactly its executed latent states, repacks these variable-length prefixes, and processes them with explicit attention masks and position IDs before decoding the answer. This procedure avoids further latent updates for examples that have already reached their budgets, while allowing examples with heterogeneous budgets to proceed through answer decoding in a batch.

\section{Experimental Setup}
\label{sec:experimental-setup}

\paragraph{Data.}
We use \texttt{whynlp/gsm8k-aug} and \texttt{whynlp/gsm8k-aug-nl}, two aligned versions of GSM8K-Aug \cite{deng2024internalize,wu2025pccot}. Each provides 385,620 training examples, 500 validation examples, and 1,319 test examples; the test split follows the official GSM8K test set \cite{cobbe2021gsm8k}. For each run, we train on the corresponding training split, select the best checkpoint by validation exact-match accuracy, and report its performance on the test split.

For OOD evaluation, we use SVAMP \cite{patel2021svamp}, GSM-Hard \cite{gao2023pal}, and MultiArith \cite{roy2015multiarith}. We concatenate the 700-example SVAMP training split and its 300-example test split, use all 1,319 GSM-Hard examples, and use the 180-example MultiArith test split. Dataset fields and representative examples are provided in Appendix~\ref{app:datasets}.

\paragraph{Difficulty scores.}
For the main experiments, we use difficulty scores assigned by the DeepSeek model family from the question alone \cite{deepseekai2024v3}, ranging from 1 (nearly one-step arithmetic) to 5 (unusually tricky). In the ablations, we also evaluate continuation negative log-likelihood (NLL) as an alternative signal for difficulty supervision. Other possible signals include the number of annotated reasoning steps and statistics of the arithmetic operators appearing in those steps. Each of these signals can define relative neighborhoods for the contrastive objective without specifying a target budget.

\paragraph{Instantiation.}
Dynamic K-Head is not tied to a particular latent-reasoning framework: it retains the framework's original training objective as $\mathcal{L}_{\mathrm{base}}$ and augments it with the difficulty-supervision loss $\mathcal{L}_K$ in Eq.~\ref{eq:objective}. In the present experiments, we instantiate the method with CoDI \cite{shen2025codi}, using GPT-2 \cite{radford2019gpt2} and Llama 3.2 1B Instruct \cite{meta2024llama32} as base models. CoDI trains a shared model through an explicit-rationale path and a recurrent latent path: the explicit path supplies hidden-state targets, and the latent path learns to compress the reasoning trajectory through self-distillation. CoDI-specific settings otherwise follow the original method.

\paragraph{Dynamic K-Head instantiation.}
We implement $g_\phi$ as one linear layer with bias and use $\mathcal{K}=\kset$. The warm-up fraction is $\rho=0.1$, with anchor $K_a=6$; the anchor-mixture weight is fixed at $\alpha=0.3$. After warm-up, $\tau$ decreases linearly from $1.5$ to $0.7$, while $\epsilon$ decreases from $0.3$ to $0.1$. Difficulty pairs are constructed within each batch using $n_+=2$ and $n_-=8$. Unless otherwise stated, we use $\lambda_K=0.03$ and sweep its value in the later ablation (Section~\ref{sec:lambda-ablation}).

\paragraph{Implementation and hardware configuration.}
Both GPT-2 and Llama experiments use LoRA fine-tuning \cite{hu2022lora}, and training and evaluation are implemented with PyTorch and Hugging Face Transformers \cite{paszke2019pytorch,wolf2020transformers}. All experiments run on NVIDIA H100 80GB HBM3 Tensor Core GPUs, with two GPUs allocated to each run. The eight-GPU server provides NVLink connectivity between every pair of GPUs.

\section{Results}

\subsection{Overall Accuracy--Cost Results}

Table~\ref{tab:main} compares No-CoT SFT, CoT SFT, Coconut \cite{hao2024coconut}, CoDI, Adaptive Latent Reasoning \cite{ning2025adaptive}, and Dynamic K-Head across GPT-2 and Llama 3.2 1B on GSM8K and GSM8K-NL. All Dynamic K-Head models use DeepSeek difficulty supervision. Relative to the fixed-budget CoDI baseline, Dynamic K-Head is consistently more accurate on Llama: accuracy increases from 52.31\% to 54.44\% on GSM8K and from 46.70\% to 50.34\% on GSM8K-NL. On GSM8K, this gain is accompanied by a 20.9\% cost reduction, from 15.21 to 12.03 tokens. On GSM8K-NL, the higher accuracy costs 1.95 additional tokens. These results show that learned pre-rollout budget selection can improve the fixed-budget latent reasoner rather than merely shorten its rollout.

\begin{figure}[t]
\centering
\includegraphics[width=\textwidth]{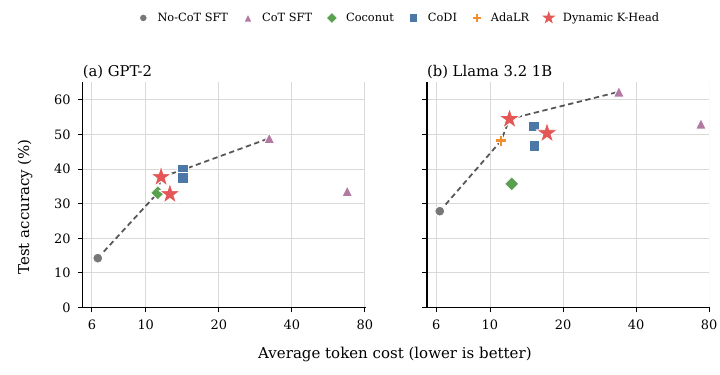}
\caption{Accuracy versus average token cost for the Table~\ref{tab:main} results, with GPT-2 on the left and Llama 3.2 1B on the right. GSM8K and GSM8K-NL are overlaid using the same method-specific markers, and the horizontal axis uses a logarithmic scale. Dashed lines indicate the Pareto frontiers.}
\label{fig:main-accuracy-cost}
\end{figure}

Among the latent methods evaluated here, Dynamic K-Head is also the closest to CoT SFT on Llama. On GSM8K, it is 7.88 accuracy points below CoT SFT while reducing average token cost from 33.96 to 12.03, a $2.82\times$ compression. On GSM8K-NL, the gap narrows to 2.73 points while cost falls from 73.95 to 17.17 tokens, a $4.31\times$ compression. Dynamic K-Head therefore recovers much of explicit CoT's answer quality with substantially less autoregressive output. Figure~\ref{fig:main-accuracy-cost} visualizes the same comparison in the accuracy--cost plane, combining both reasoning formats within each backbone panel. Dynamic K-Head contributes a point to the overall Pareto frontier for both backbones.

\begin{table}[t]
\caption{Accuracy and average token cost on the official GSM8K test set across two backbones and the GSM8K and GSM8K-NL training variants. AdaLR denotes Adaptive Latent Reasoning. All Dynamic K-Head rows use DeepSeek difficulty supervision. Token cost follows Eq.~\ref{eq:cost}. Within each column, bold and underlined values indicate the best and second-best results, respectively; CoT SFT is excluded from the accuracy ranking, and No-CoT SFT is excluded from the cost ranking.}
\label{tab:main}
\centering
\scriptsize
\resizebox{\textwidth}{!}{\begin{tabular}{lrrrr}
\toprule
& \multicolumn{2}{c}{GSM8K} & \multicolumn{2}{c}{GSM8K-NL} \\
\cmidrule(lr){2-3}\cmidrule(lr){4-5}
Method & Accuracy (\%) & Avg. token cost & Accuracy (\%) & Avg. token cost \\
\midrule
\multicolumn{5}{l}{\textbf{GPT-2}} \\
No-CoT SFT & 14.25 & 6.34 & 14.25 & 6.34 \\
CoT SFT & 48.90 & 32.27 & 33.59 & 67.60 \\
Coconut & 33.13 & \textbf{11.21} & -- & -- \\
CoDI & \textbf{39.80} & 14.24 & \textbf{37.38} & \underline{14.26} \\
Dynamic K-Head & \underline{37.68} & \underline{11.56} & \underline{32.75} & \textbf{12.57} \\
\midrule
\multicolumn{5}{l}{\textbf{Llama 3.2 1B}} \\
No-CoT SFT & 27.82 & 6.21 & 27.82 & 6.21 \\
CoT SFT & 62.32 & 33.96 & 53.07 & 73.95 \\
Coconut & 35.71 & 12.29 & -- & -- \\
CoDI & \underline{52.31} & 15.21 & \underline{46.70} & \textbf{15.22} \\
AdaLR & 48.14 & \textbf{11.09} & -- & -- \\
Dynamic K-Head & \textbf{54.44} & \underline{12.03} & \textbf{50.34} & \underline{17.17} \\
\bottomrule
\end{tabular}
}
\end{table}

\subsection{Out-of-Distribution Evaluation}

\begin{figure}[!htbp]
\centering
\includegraphics[width=\textwidth]{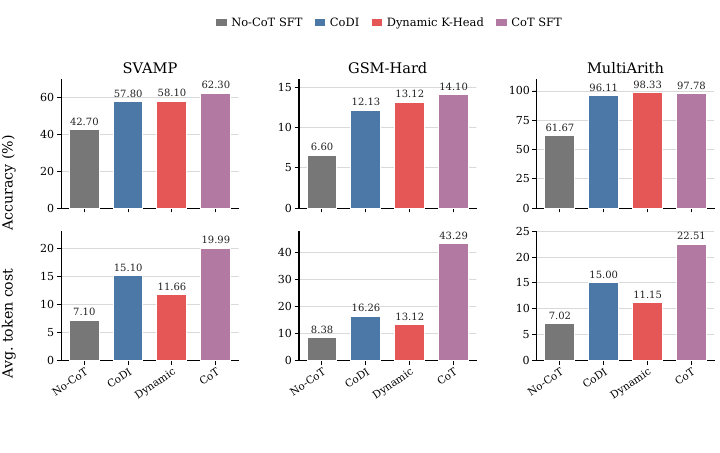}
\caption{Accuracy (top) and average token cost (bottom) on three OOD datasets. Within each column, methods are ordered as No-CoT SFT, CoDI ($K=6$), Dynamic K-Head, and CoT SFT. Each panel uses a dataset-specific vertical scale.}
\label{fig:ood-accuracy-cost}
\end{figure}

We evaluate the Llama 3.2 1B models on the three arithmetic OOD datasets described in Section~\ref{sec:experimental-setup}. Figure~\ref{fig:ood-accuracy-cost} compares Dynamic K-Head with CoT SFT, No-CoT SFT, and fixed-$K$ CoDI checkpoints. Token cost follows Eq.~\ref{eq:cost}.

Compared with the fixed-$K$ CoDI baseline, Dynamic K-Head reduces average cost by 19.3--25.7\% across the three datasets while improving accuracy by 0.30--2.22 points. This joint improvement indicates that instance-dependent budget selection transfers beyond the training distribution rather than merely shortening in-domain rollouts. Relative to CoT SFT, Dynamic K-Head reduces cost by 41.7--69.7\%, corresponding to $1.71\times$--$3.30\times$ compression. CoT SFT is 4.20 and 0.99 points more accurate on SVAMP and GSM-Hard, respectively; on MultiArith, Dynamic K-Head is 0.56 points more accurate while using half the cost.

Dynamic K-Head remains highly accurate on MultiArith but reaches only 13.12\% on GSM-Hard. The budget selector averages 2.15--2.74 latent updates across the three datasets and never selects $K=8$, suggesting that the remaining errors reflect limited coverage of the shifted arithmetic demands rather than exhaustion of the available budget.

\FloatBarrier
\subsection{Online vs. External Difficulty Supervision}

The key question is whether budget selection benefits more from the student's evolving notion of difficulty or from a fixed external ordering. Table~\ref{tab:difficulty-supervision} compares fixed-budget CoDI with Dynamic K-Head trained using continuation NLL or DeepSeek difficulty scores computed from questions alone.

\begin{table}[t]
\caption{Fixed-budget CoDI and Dynamic K-Head with two signals for difficulty supervision on GSM8K. NLL denotes online continuation NLL recomputed at each training epoch; difficulty denotes fixed DeepSeek scores computed from questions alone. Within each backbone block, bold and underlined values indicate the best and second-best results, respectively.}
\label{tab:difficulty-supervision}
\centering
\small
\begin{tabular}{lrr}
\toprule
Method & Accuracy (\%) & Avg. token cost \\
\midrule
\multicolumn{3}{l}{\textbf{GPT-2}} \\
CoDI & \textbf{39.80} & \underline{14.24} \\
Dynamic K-Head (NLL) & \underline{38.29} & 14.37 \\
Dynamic K-Head (difficulty) & 37.68 & \textbf{11.56} \\
\midrule
\multicolumn{3}{l}{\textbf{Llama 3.2 1B}} \\
CoDI & 52.31 & 15.21 \\
Dynamic K-Head (NLL) & \underline{53.15} & \underline{14.65} \\
Dynamic K-Head (difficulty) & \textbf{54.44} & \textbf{12.03} \\
\bottomrule
\end{tabular}

\end{table}

On Llama, DeepSeek difficulty supervision reaches 54.44\% at 12.03 tokens and improves both metrics over CoDI and continuation NLL. On GPT-2, it lowers cost to 11.56 tokens but also lowers accuracy to 37.68\%. This weaker result may partly reflect GPT-2's substantially smaller capacity, which can limit both the quality of $h_q$ for budget selection and the reasoner's ability to preserve accuracy under shorter computation.

The signals also differ in how they evolve. Continuation NLL is an online, student-dependent signal recomputed with the current model at every training epoch; it therefore tracks the student's changing uncertainty. DeepSeek difficulty is a fixed, external question-level signal that provides a stable ordering throughout training. Since $\mathcal{L}_K$ uses these values to construct relative neighborhoods rather than target budgets, the choice between an adaptive internal signal and a stable external signal changes the representations learned by the K-head. The results indicate that neither notion of difficulty defines a universal compute target: its value depends on the capacity and representations of the student model.

\subsection{Difficulty Supervision Controls Budget Selection}
\label{sec:lambda-ablation}

Table~\ref{tab:ablation} reveals distinct patterns of budget selection. At $\lambda_K=0$, $H(K\mid Q)=1.715$ nearly equals $H(K)=1.781$ and $I(Q;K)=0.066$, indicating diffuse, nearly question-independent budget distributions.

\begin{table}[!htbp]
\caption{Accuracy, average token cost, marginal entropy $H(K)$, predictive entropy $H(K\mid Q)$, and question--budget mutual information $I(Q;K)$ for the loss-weight ablation on GSM8K. All information quantities are measured in nats.}
\label{tab:ablation}
\centering
\scriptsize
\begin{tabular}{rrrrrr}
\toprule
$\lambda_K$ & Accuracy (\%) & Avg. cost & $H(K)$ & $H(K\mid Q)$ & $I(Q;K)$ \\
\midrule
0.00 & 50.87 & 13.11 & 1.781 & 1.715 & 0.066 \\
0.01 & 54.51 & 13.33 & 0.441 & 0.131 & 0.310 \\
0.03 & 54.44 & 12.03 & 0.540 & 0.097 & 0.443 \\
0.05 & 52.62 & 10.27 & 0.122 & 0.006 & 0.115 \\
0.10 & 49.05 & 10.89 & 0.724 & 0.300 & 0.424 \\
\bottomrule
\end{tabular}

\end{table}

Moderate difficulty supervision produces decisive, question-dependent budget selection. Although $\lambda_K=0.01$ is most accurate (54.51\%), 0.03 nearly matches it (54.44\%) at lower cost (12.03 versus 13.33 tokens) and pairs low predictive entropy (0.097) with the highest mutual information (0.443), giving the strongest observed balance.

Higher weights separate budget concentration from utility. At 0.05, $H(K\mid Q)=0.006$ and $H(K)=0.122$ indicate near-global concentration; its 52.62\% accuracy at the lowest cost (10.27 tokens) makes it cheaper but less adaptive. At 0.10, mutual information is 0.424 but accuracy falls to 49.05\%.

The low-score train--evaluation shift ($\leq 2$: 91.3\% versus 71.8\%/71.9\%; Appendix~\ref{app:difficulty}) may contribute to high-weight concentration.

\FloatBarrier

\section{Limitations and Broader Considerations}
\label{sec:limitations}

\paragraph{Small models.}
Our experiments are limited to GPT-2 and Llama, so whether the findings generalize to other model families and larger scales remains unknown.

\paragraph{External difficulty scores.}
Teacher-assigned scores may contain systematic biases and are not guaranteed to accurately reflect the task difficulty experienced by the student model. Future work should compare multiple scoring models.

\paragraph{Weak supervision.}
The contrastive objective provides only weak, question-level signals for budget selection. Because the K-head reads a representation produced by the shared backbone, it may indirectly inherit some outcome-related information while the backbone is optimized by $\mathcal{L}_{\mathrm{base}}$ under sampled budgets. This pathway nevertheless provides no explicit per-example credit assignment identifying which candidate $K$ values would yield a correct answer, nor counterfactual evidence for budgets that were not executed. The learned K-head representations may therefore align with difficulty-score neighborhoods without matching the reasoner's actual budget-specific success pattern. Stronger supervision could use outcomes collected across multiple budgets for each question, at the cost of additional rollouts.

\section{Conclusion}

We introduced \method{}, a pre-rollout budget selector that treats latent budgets as non-monotonic, instance-conditioned choices and learns them with difficulty supervision. We also proposed end-to-end cost accounting that includes latent updates, special tokens, and decoded text. With CoDI on Llama 3.2 1B, \method{} reaches 54.44\% accuracy on GSM8K at 12.03 average tokens, improving fixed-$K=6$ CoDI by 2.12 points while reducing cost by 20.9\%, and it retains joint accuracy--cost gains on three out-of-distribution datasets. Budget-selection quality depends on the granularity and balance of difficulty coverage and on how directly supervision reflects budget utility. Future work should therefore combine finer-grained, better-balanced training data with more direct outcome-based supervision.

\begin{credits}
\subsubsection{Code, Data, and Model Availability.}
We will open-source the code and publicly release the difficulty-annotated datasets and model weights. These resources are currently being organized and prepared for release.

\subsubsection{Declaration of Generative AI Use.}
During manuscript preparation, the authors used OpenAI GPT-5.6-sol solely to improve readability, clarity, grammar, and style. The authors reviewed and approved all edits and take responsibility for the final content.

\subsubsection{\discintname}
The authors have no competing interests to declare that are relevant to the content of this article.
\end{credits}

\bibliographystyle{splncs04}
\bibliography{references}

\clearpage
\appendix
\section{Dataset Formats and Examples}
\label{app:datasets}

\subsection{GSM8K-Aug Training Data}

Both \texttt{whynlp/gsm8k-aug} and \texttt{whynlp/gsm8k-aug-nl} store each record with three fields: \texttt{question} is a string, \texttt{steps} is a list of strings, and \texttt{answer} is a string. The datasets align questions and answers but serialize \texttt{steps} differently. In \texttt{whynlp/gsm8k-aug}, each step is a compact calculator-style expression enclosed by double angle brackets. In \texttt{whynlp/gsm8k-aug-nl}, each step is a natural-language sentence.

\paragraph{Aligned example.}
The following record occurs in both formats.

\begin{quote}
\small
\textbf{Question:} Out of 600 employees in a company, 30\% got promoted while 10\% received a bonus. How many employees did not get either a promotion or a bonus?

\textbf{\texttt{whynlp/gsm8k-aug} steps:}
\begin{enumerate}
    \item \verb|<<600*30/100=180>>|
    \item \verb|<<600*10/100=60>>|
    \item \verb|<<180+60=240>>|
    \item \verb|<<600-240=360>>|
\end{enumerate}

\textbf{\texttt{whynlp/gsm8k-aug-nl} steps:}
\begin{enumerate}
    \item $600 \times 30/100 = 180$ employees were promoted.
    \item $600 \times 10/100 = 60$ employees received a bonus.
    \item A total of $180+60=240$ employees received a promotion or a bonus.
    \item Therefore, $600-240=360$ employees did not get either a promotion or a bonus.
\end{enumerate}

\textbf{Answer:} \texttt{360}
\end{quote}

\subsection{OOD Evaluation Data}

\noindent\textbf{SVAMP.} We concatenate the 700-example training split and the 300-example test split from \path{ChilleD/SVAMP}; the model input is \texttt{question\_concat}, and the gold answer is \texttt{Answer}.

\noindent\textbf{GSM-Hard.} We use all 1,319 examples from \path{juyoung-trl/gsm-hard}, with \texttt{instruction} as the input and \texttt{response} as the answer.

\noindent\textbf{MultiArith.} We use the 180-example test split from \path{ChilleD/MultiArith}, with fields \texttt{question} and \texttt{final\_ans}. Representative records from the evaluated splits follow; surrounding whitespace is omitted.

\paragraph{SVAMP example (test split).}
\begin{quote}
\small
\textbf{Question (\texttt{question\_concat}):} There were 106 dollars in Olivia's wallet. After she visited a supermarket and a showroom there were 26 dollars left. If she spent 49 dollars at the showroom How much did she spend at the supermarket?

\textbf{Answer (\texttt{Answer}):} \texttt{31}
\end{quote}

\paragraph{GSM-Hard example.}
\begin{quote}
\small
\textbf{Question (\texttt{instruction}):} A robe takes 2287720 bolts of blue fiber and half that much white fiber. How many bolts in total does it take?

\textbf{Answer (\texttt{response}):} \texttt{3431580.0}
\end{quote}

\paragraph{MultiArith example (test split).}
\begin{quote}
\small
\textbf{Question (\texttt{question}):} Paige had 11 songs on her mp3 player. If she deleted 9 old songs from it and then added 8 new songs, how many songs does she have on her mp3 player?

\textbf{Answer (\texttt{final\_ans}):} \texttt{10}
\end{quote}

\section{A Small Latent Budget Can Still Produce a Long Continuation}
\label{app:continuation}

The following is a real, correctly answered example from an official GSM8K test evaluation of a GPT-2 Dynamic K-Head model. The selector chose only one latent update, $\widehat K=1$, but the model then continued the computation in explicit text.

\begin{quote}
\small
\textbf{Question:} Tom plants 10 trees a year. Every year he also chops down 2 trees a year. He starts with 50 trees. After 10 years 30\% of the trees die. How many trees does he have left?

\textbf{Selected latent budget:} $\widehat K=1$

\textbf{Decoded continuation:}

\verb|8*10=80>> <<50+80=130>>|\\
\verb|<<130*30/100=39>>The answer is: 91|

\textbf{Gold answer:} \texttt{91}
\end{quote}

The continuation contains 28 generated tokens, including three explicit calculation steps and the final answer. Counting the \texttt{BOT} and \texttt{EOT} tokens used by this CoDI execution, the example costs $1+2+28=31$ tokens despite using only one latent update. This illustrates why $K$ alone is not a reliable proxy for end-to-end reasoning/output cost.

\section{Difficulty-Score Distributions}
\label{app:difficulty}

The DeepSeek scores, computed from questions alone, are more concentrated in the augmented training set than in the evaluation sets. Scores at most 2 account for 91.3\% of training questions, compared with 71.8\% of validation and 71.9\% of test questions. Validation and test have closely matched profiles, while both place more mass than training at scores 2.5 and 3.5.

\begin{figure}[t]
\centering
\includegraphics[width=\textwidth]{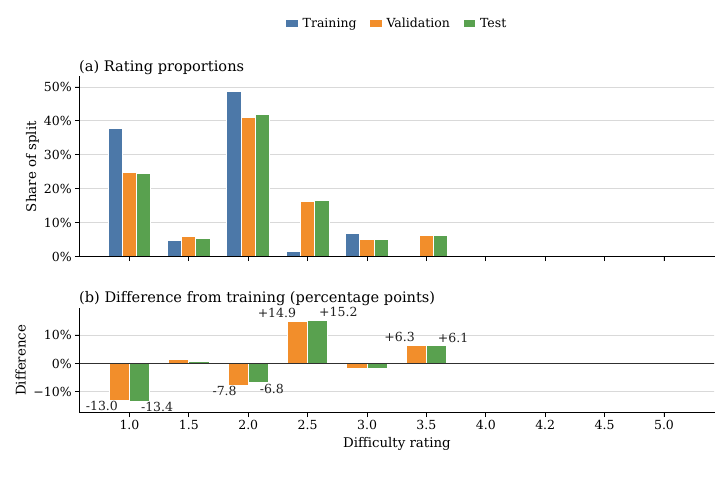}
\caption{DeepSeek difficulty scores, computed from questions alone, across training, validation, and test. The lower panel reports validation-minus-training and test-minus-training differences in percentage points.}
\end{figure}

\section{Information Quantities for Budget Selection}
\label{app:information}

Let $Q$ be uniformly distributed over the $N=1{,}319$ GSM8K test questions, and let $p_i(k)=p_\phi(k\mid q_i;\tau=1)$ be the soft K-head distribution for question $q_i$. We compute
\begin{align*}
H(K\mid Q) &= \frac{1}{N}\sum_{i=1}^{N}-\sum_{k\in\mathcal{K}}p_i(k)\ln p_i(k), \\
\bar p(k) &= \frac{1}{N}\sum_{i=1}^{N}p_i(k), \\
H(K) &= -\sum_{k\in\mathcal{K}}\bar p(k)\ln\bar p(k), \\
I(Q;K) &= H(K)-H(K\mid Q).
\end{align*}
Here $H(K\mid Q)$ is the mean per-question predictive entropy, $H(K)$ is the entropy of the aggregate soft-budget distribution, and $I(Q;K)$ measures how strongly the soft budget distribution varies across questions. All quantities use natural logarithms and are reported in nats.

\end{document}